# Hybrid clustering-classification neural network in the medical diagnostics of reactive arthritis

**Yevgeniy Bodyanskiy, Olena Vynokurova**

Control Systems Research Laboratory, Kharkiv National University of Radio Electronics, Kharkiv, 61166, Ukraine

Email: {yevgeniy.bodyanskiy, olena.vynokurova}@nure.ua

**Volodymyr Savvo, Tatiana Tverdokhlib**

Pediatrics Department, Kharkiv Medical Academia of Post-Graduate Education, Kharkiv, 61176, Ukraine

Email: savvovm50@gmail.com, tanyamar82@gmail.com

**Pavlo Mulesa**

Cybernetics and Applied Mathematics Department, Uzhhorod National University, Uzhhorod, 88000, Ukraine

Email: ppmulesa@gmail.com

*Abstract*— In the paper, the hybrid clustering-classification neural network is proposed. This network allows to increase a quality of information processing under the condition of overlapping classes due to the rational choice of learning rate parameter and introducing special procedure of fuzzy reasoning in the clustering-classification process, which occurs both with external learning signal ("supervised"), and without one ("unsupervised"). As similarity measure neighborhood function or membership one, cosine structures are used, which allow to provide a high flexibility due to self-learning-learning process and to provide some new useful properties. Many realized experiments have confirmed the efficiency of proposed hybrid clustering-classification neural network; also, this network was used for solving diagnostics task of reactive arthritis.

*Index Terms*— hybrid clustering-classification neural network, supervised/unsupervised learning, overlapping classes, diagnostics, reactive arthritis.

I. INTRODUCTION

Self-organizing maps (SOM) and neural networks of learning vector quantization (LVQ) have seen extensive use for solving different problems in Data Mining domain (clustering, classification, fault detection and compression of information etc.). This type of neural networks was proposed by T. Kohonen [1, 2] and represents, in fact, a single-layer feedforward architecture, which provides an operator for mapping of input space into the output space. Operation-wise SOM and LVQ are quite similar to each neuron is fed input signal (sample) producing output, which is used during competition stage to determine winning neuron – usually the one with maximum output signal value. Vector of synaptic weights for winning neuron is the one closest to the input sample in terms of the metric chosen (which is Euclidian metric in most cases). Next is neurons adjustment phase. Synaptic weights of the winning neuron gets moved closer to input sample. Alternatively, a subset of neurons (rather than a single one) can be adjusted – those determined to be "reasonably close" to the input sample are updated. Resulting network output, however, is determined exclusively by winning neuron (this principle is usually referred to as "Winner-Takes-All" (WTA)). It is this principle (WTA) which negatively affects accuracy in case when there are overlapping clusters in underlying data.

Taking into account the above mentioned properties of SOM and LVQ networks, it makes sense to introduce fuzzy classification capabilities on top of them, while preserving online operation. In [8, 9] fuzzy self-organizing map was proposed, in which conventional neurons are replaced by fuzzy rules. This neural network shows enough high efficiency, but its learning properties were significantly lost especially in on-line mode. In [5, 10, 11] fuzzy clustering Kohonen network and fuzzy linear vector quantization network are described. In fact, such networks are neural representation of fuzzy c-means (FCM) [3], which is far enough from SOM and LVQ mathematical tool and designed for operation in batch mode.

II. PROBLEM STATEMENT

Let us consider single-layer neural network with lateral connections containing receptors and neurons in the Kohonen layer with each neuron being characterized by a vector of its synaptic weights. During learning stage input vector is fed to the inputs of all neurons (usually adaptive linear associators) (here - either the number of observation in a table "object-properties", or current discrete time for on-line processing mode) and neurons produce the scalar signals on their outputs

$$y_j(k) = w_j^T(k)x(k), \quad j = 1, 2, ..., m. \quad (1)$$

Note that neuron's output depends on current values of synaptic weights vector, assuming iterative learning algorithm.

Each input vector can activate either single neuron ($w_j$) or a set of neighboring neurons – this also depends on learning algorithm chosen.

Self-organization procedure is based on the competitive learning approach (self-learning) and begins with the initialization of synaptic weights. Selecting initial values for weights from a uniform random distribution over input space has become a common practice. In addition, weights are usually normalized to reside on unit hypersphere:

$$\|w_j(k)\|^2 = w_j^T(k)w_j(k) = 1. \quad (2)$$

Self-organizing usually contains three stages [2]: competition, cooperation, and synaptic adaptation. This paper introduces additional stage to this process – fuzzy inference, which allows an online learning algorithm to classify data samples belonging to several classes simultaneously.

### III. LEARNING ALGORITHM FOR SOM

The competition process is started to analysis of current pattern $x(k)$, which is fed to all neurons of Kohonen's layer from receptive (zero) layer. For each neuron the distance between input sample and a vector of synaptic weights is computed:

$$D(w_j(k), x(k)) = \|x(k) - w_j(k)\|. \quad (3)$$

In case when all inputs and synaptic weights are normalized according to (2), i.e.

$$\|w_j(k)\| = \|x(k)\| = 1, \quad (4)$$

and Euclidian metric is used, the proximity measure between the vectors $w_j(k)$ and $x(k)$ can be written in the following way:

$$w_j^T(k)x(k) = y_j(k) = \cos(w_j(k), x(k)) = \cos\theta_j(k). \quad (5)$$

So the expression (3) takes the form

$$D(w_j(k), x(k)) = \sqrt{2(1 - y_j(k))}. \quad (6)$$

Using relation (5), we can replace metric (3) with a simpler construction, usually referred to as a measure of similarity [12]

$$\psi(w_j(k), x(k)) = \max\{0, \cos\theta_j(k)\}, \quad (7)$$

which satisfies softer conditions compared to classic measure requirements:

$$\psi(w_j(k), x(k)) \geq 0,$$
$$\psi(w_j(k), w_j(k)) = 1, \quad (8)$$
$$\psi(w_j(k), x(k)) = \psi\{x(k), w_j(k)\}.$$

Next we look for a winning neuron that has a biggest value of similarity to the input vector in the sense (6) or (7), i. e.

$$D(w_j^*(k), x(k)) = \min_l D(w_l(k), x(k)) \quad (9)$$

or

$$\psi(w_j^*(k), x(k)) = \max_l \psi(w_l(k), x(k)). \quad (10)$$

After that, we tune values of synaptic weights using WTA self-learning rule in form

$$w_j(k+1) = \begin{cases} w_j^*(k) + \eta(k)(x(k) - w_j^*(k)), \\ \quad \text{if } j\text{-th neuron won in} \\ \quad \text{the competition,} \\ w_j(k) \quad \text{otherwise.} \end{cases} \quad (11)$$

In a nutshell, expression (11) means that synaptic weights vector of winner is moved closer to the input pattern $x(k)$ by a value which is proportional to learning rate $0 < \eta(k) < 1$.

Choice of learning rate value $\eta(k)$ defines the convergence rate of self-learning process, and is usually based on empirical reasons, at that the parameter must monotonically decrease in on time.

It is easy to see, that first relation of the rule (11) minimizes the criterion

$$E_j^k = \sum_{\tau=1}^{k_j} \|x(\tau) - w_j\|^2 \quad (12)$$

(here $k_j$ is a number of data in the sampling with dimensions $k$, when $j$-th neuron was winner), notably in practice as synaptic weights estimation conventional arithmetic mean is used:

$$w_j(k) = \frac{1}{k_j} \sum_{\tau=1}^{k_j} x(\tau). \quad (13)$$

Therefore, in fact, self-learning rule (3) is stochastic approximation procedure [13], and learning rate parameter $\eta(k)$ must be selected according to the conditions of A. Dvoretzky [14]. It is known that using the following value

$$\eta(k) = \frac{1}{k_j} \quad (14)$$

leads to slowdown of the learning process.

Requirement of monotonic decreasing of the parameter $\eta(k)$ leads to expression in form

$$\eta(k) = r^{-1}(k),$$
$$r(k) = \alpha r(k-1) + \|x(k)\|^2, \ 0 \leq \alpha \leq 1. \quad (15)$$

When taking into account normalization to unit hypersphere (4) we have:

$$r(k) = \alpha r(k-1) + 1, \quad (16)$$

which with $\alpha = 1$ gives a well-known expression, that was introduced in [15].

Changing the forgetting parameter $\alpha$ provides enough variance for the learning rate to both satisfy Dvoretzky condition and adjust for characteristics of different data sets:

$$\frac{1}{k_j} \leq \eta(k) \leq 1. \quad (17)$$

Note that adjusting synaptic weights with (13) in general breaks normalization (4). In order to maintain it, we should modify our learning algorithm:

$$w_j(k+1) = \begin{cases} \dfrac{w_j^*(k) + \eta(k)(x(k) - w_j^*(k))}{\|w_j^*(k) + \eta(k)(x(k) - w_j^*(k))\|}, \\ \quad \text{if } j-th \text{ neuron won in the} \\ \quad \text{competition,} \\ \eta(k) = r^{-1}(k), \\ r(k) = \alpha r(k-1) + 1, \quad 0 \leq \alpha \leq 1, \\ w_j(k), \text{ otherwise.} \end{cases} \quad (18)$$

In order to better adjust to input data, a so-called "cooperation stage" is frequently introduced to SOM learning process. During this stage, a winning neuron defines a local domain of topological neighborhood, where weights are adjusted for a set of neurons rather than only for a winning one. Those neurons closer to the winner receive a bigger adjustment.

This topological domain is defined by neighborhood function $\varphi(j,p)$, which depends on distance $D(w_j^*(k), w_p(k))$ between winner $w_j^*(k)$ and any of Kohonen's layer neurons $w_p(k)$, $p = 1, 2, \ldots, m$ and some parameter $\sigma$, which sets its width.

Usually $\varphi(j,p)$ is the bell-shaped function, symmetrical with respect to the maximum in point $w_j^*(k)$ ($\varphi(j,j) = 1$) and decreasing when distance $D(w_j^*(k), w_p(k))$ increases. Gaussian function is commonly used here:

$$\varphi(j,p) = \exp\left(-\frac{\|w_j^*(k) - w_p(k)\|^2}{2\sigma^2}\right). \quad (19)$$

Adding neighborhood function results in the following learning algorithm:

$$w_p(k+1) = w_p(k) + \eta(k)\varphi(j,p)(x(k) - w_p(k)), \quad (20)$$

which minimizes criterion

$$E_p^k = \sum_{\tau=1}^{k} \varphi(j,p) \|x(\tau) - w_p\|^2 \quad (21)$$

This criterion is commonly referred to as "Winner Takes More" (WTM).

The necessity to maintain normalization to unit hypersphere (4) leads to the expression in form

$$\begin{cases} w_p(k+1) = \dfrac{w_p(k) + \eta(k)\varphi(j,p)(x(p) - w_p(k))}{\|w_p(k) + \eta(k)\varphi(j,p)(x(p) - w_p(k))\|}, \\ p = 1, 2, \ldots, m, \\ \eta(k) = r^{-1}(k), \; r(k) = \alpha r(k-1) + 1, \; 0 \leq \alpha \leq 1. \end{cases} \quad (22)$$

It is possible to skip the entire competition by tuning synaptic weights of network depending on their similarity with the current vector-pattern $x(k)$. In this case instead of conventional neighborhood function $\varphi(j,p)$ depending on winner $w_j^*(k)$, we can use similarity measure (7), that depends on $x(k)$.

In this case instead of (20) we can use the rule in form:

$$w_p(k+1) = w_p(k) + \\ + \eta(k)\psi(w_p(k), x(k))(x(k) - w_p(k)) = \quad (23) \\ = w_p(k) + \eta(k)\max\{0, y_p(k)\}(x(k) - w_p(k)),$$

which reminds «INSTAR» learning rule of S. Grossberg [16].

For maintaining (4) the rule (23) has to be rewritten in the form

$$\begin{cases} w_p(k+1) = \dfrac{w_p(k)+\eta(k)\psi(w_p(k),x(k))(x(k)-w_p(k))}{\|w_p(k)+\eta(k)\psi(w_p(k),x(k))(x(k)-w_p(k))\|}, \\ \eta(k)=r^{-1}(k),\ r(k)=\alpha r(k-1)+1,\ 0\leq\alpha\leq 1. \end{cases} \quad (24)$$

In many real world problems clusters can overlap as shown on Fig. 1, where * denotes patterns, belonging to $j$-th clusters, and o – $p$-th ones. In this case vector $x(k)$ belongs $j$-th cluster with proportional membership level $\cos\theta_j(k)$, and with proportional level $\cos\theta_p(k)$ - to $p$-th one. Synaptic weights concentrated in crosshatched region, don't have relationship to the pattern $x(k)$ according to (7).

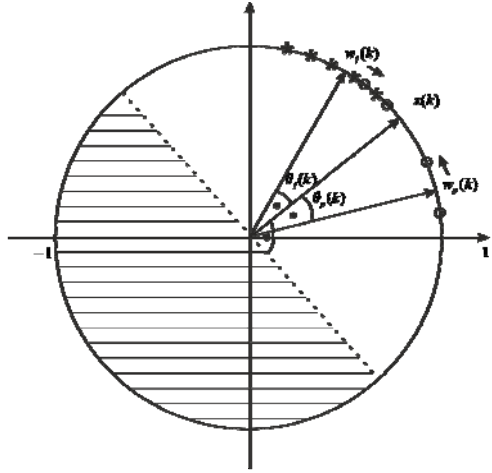

Fig. 1. The overlapping clusters

Using similarity measure, that is shown on fig.2, we can introduce the membership estimate of pattern $x(k)$ to $j$-th class in form:

$$0 \leq \mu_{w_j(k)}(x(k)) = \dfrac{\psi(w_j(k),x(k))}{\sum_{l=1}^{m}\psi(w_l(k),x(k))} \leq 1. \quad (25)$$

## IV. Learning algorithm for LVQ

Learning vector quantization neural networks in contrast to self-learning SOM adjust their parameters based on external learning (reference) signal, which fixes the membership of each pattern $x(k)$ to a particular class.

The main idea of LVQ neural network is to build a compact representation of large data set based on a restricted set of prototype samples (centroids) $w_j(k)$, $j=1,2,\ldots,m$, that provide sufficiently accurate approximation of the original space $X$.

For each input vector $x(k)$ which is normalized according to (4), we look for a winning neuron $w_j^*(k)$ that corresponds to a centroid of a certain class. In other words, winner is defined as a neuron with minimal distance to the input vector (9) or, which is the same, with maximal similarity measure (10).

Since the learning is supervised, membership of the vector $x(k)$ to specific domain $X_j$ of the space $X$ is known, which allows considering two typical situations, which occur in the vector quantization:

- the input vector $x(k)$ and neuron-winner $w_j^*(k)$ belong to the same cell of Voronoy [2];

- the input vector $x(k)$ and neuron-winner $w_j^*(k)$ belong to the different cells of Voronoy.

Than corresponding learning LVQ-rule can be written in form:

$$w_j(k+1) = \begin{cases} w_j^*(k) + \eta(k)(x(k) - w_j^*(k)), \\ \text{if } x(k) \text{ and } w_j^*(k) \text{ belong to the same cell,} \\ \\ w_j^*(k) - \eta(k)(x(k) - w_j^*(k)), \\ \text{if } x(k) \text{ and } w_j^*(k) \text{ belong to the different cells,} \\ \\ w_j(k) \text{ for neurons,} \\ \text{which aren't won in instant } k. \end{cases} \quad (26)$$

The rule (26) has a clear physical interpretation: if the winning neuron and presented sample belong to the same class, than prototype $w_j^*(k)$ is moved to $x(k)$; otherwise prototype $w_j^*(k)$ is pushed away from $x(k)$, effectively increasing the distance $D(w_j^*(k), x(k))$, i.e. solving the maximization task based on criterion (12).

As for the choice of learning rate parameter $\eta(k)$, that common recommendations are the same that for SOM, i.e. the learning rate parameter must monotonically decrease in process controlled learning.

In [17] it was proved that LVQ tuning algorithm converges in case of learning rate $\eta(k)$ satisfies Dvoretzky conditions. This, in turn, allows choosing $\eta(k)$ according to Goodwin-Ramadge-Caines algorithm [13], or in the previously defined form (16) with $\alpha = 1$, which is essentially the same. When $\alpha < 1$, the algorithm gets tracking properties and handles the case when class centroids are drifting.

For providing to fulfillment of the condition (4) it possible to introduce modification of rule (26) in the form (18):

$$w_j(k+1) = \begin{cases} \dfrac{w_j^*(k) + \eta(k)(x(k) - w_j^*(k))}{\left\| w_j^*(k) + \eta(k)(x(k) - w_j^*(k)) \right\|}, \\ \text{if } x(k) \text{ and } w_j^*(k) \text{ belong to the same cell,} \\ \\ \dfrac{w_j^*(k) - \eta(k)(x(k) - w_j^*(k))}{\left\| w_j^*(k) - \eta(k)(x(k) - w_j^*(k)) \right\|}, \\ \text{if } x(k) \text{ and } w_j^*(k) \text{ belong to the different cells,} \\ \\ w_j(k) \text{ for neurons, which are not won in instant } k. \end{cases} \quad (27)$$

First and third expressions in formula (27) are completely identical to WTA – self-learning algorithm, which the second one represents a "push-back" scenario.

Let's consider a situation, shown in Fig. 2, when neuron $w_j^*(k)$ won the competition. At the same time current sample $x(k)$ belongs to a class with different centroid $w_p(k)$. Now we need to «push» vector $w_j^*(k)$ away so, that $x(k)$ was equidistant from $w_j^*(k)$ and from $w_p(k)$.

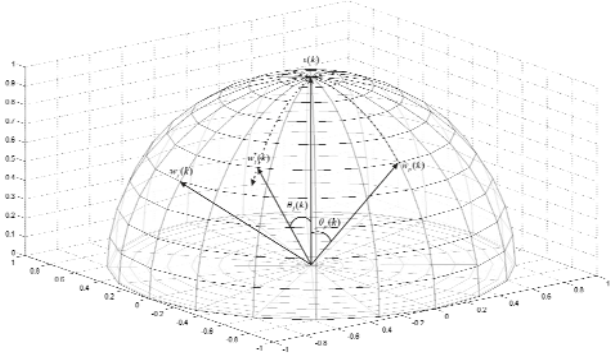

Fig. 2. Learning of LVQ

Applying elementary transformations, we get the following formulas:

$$w_j(k+1) = w_j^*(k) - \eta(k)(x(k) - w_j^*(k)), \quad (28)$$

$$\begin{aligned}x^T(k)w_j(k+1) = \\ = x^T(k)w_j^*(k) - \eta(k)\|x(k)\|^2 - \eta(k)x^T(k)w_j^*(k)\end{aligned}, \quad (29)$$

$$\begin{aligned}\cos(w_j(k+1), x(k)) = \\ = \cos(w_j^*(k), x(k)) - \eta(k)(1 + \cos(w_j^*(k), x(k)))\end{aligned} \quad (30)$$

In order to satisfy

$$\cos(w_j(k+1), x(k)) = \cos(w_p(k), x(k)) \quad (31)$$

we need to set $\eta(k)$ in form

$$\eta(k) = \frac{\cos(w_j^*(k), x(k)) - \cos(w_p(k), x(k))}{\cos(w_j^*(k), x(k)) + 1} = \\ = \frac{x^T(k)w_j^*(k) - x^T(k)w_p(k)}{x^T(k)w_j^*(k) - x^T(k)x(k)}. \quad (32)$$

After one step of the weights tuning the pattern $x(k)$ will belong equally to both $w_j(k+1)$ and $w_p(k) = w_p(k+1)$, i.e.

$$\mu_{w_j(k+1)}(x(k)) = \mu_{w_p(k+1)}(x(k)) = 0,5. \quad (33)$$

In case when several classes are overlapping, we can use estimate (25) for computing membership levels.

V. HYBRID CLUSTERING-CLASSIFICATION NEURAL NETWORK WITH SUPERVIZED/ UNSUPERVISED LEARING

A promising application of Kohonen neural network is adaptive pattern recognition, which implemented by the systems. The implementation consists of sequentially connected SOM and LVQ [2] layers. First part of network operates in self-learning mode, while the second one adds supervised component to the process.

From an input vector $x(k)$ which is fed to the input system, SOM network extracts a feature sample $y(k)$ in a space with sufficiently reduced dimensionality. This simplifies a problem in hand without significant loss of information.

On the second stage LVQ is trained to classify of incoming pattern $y(k)$ using supervised learning. A distinctive feature of proposed system is the connection of two identical architectures, that are tuned by different learning algorithms.

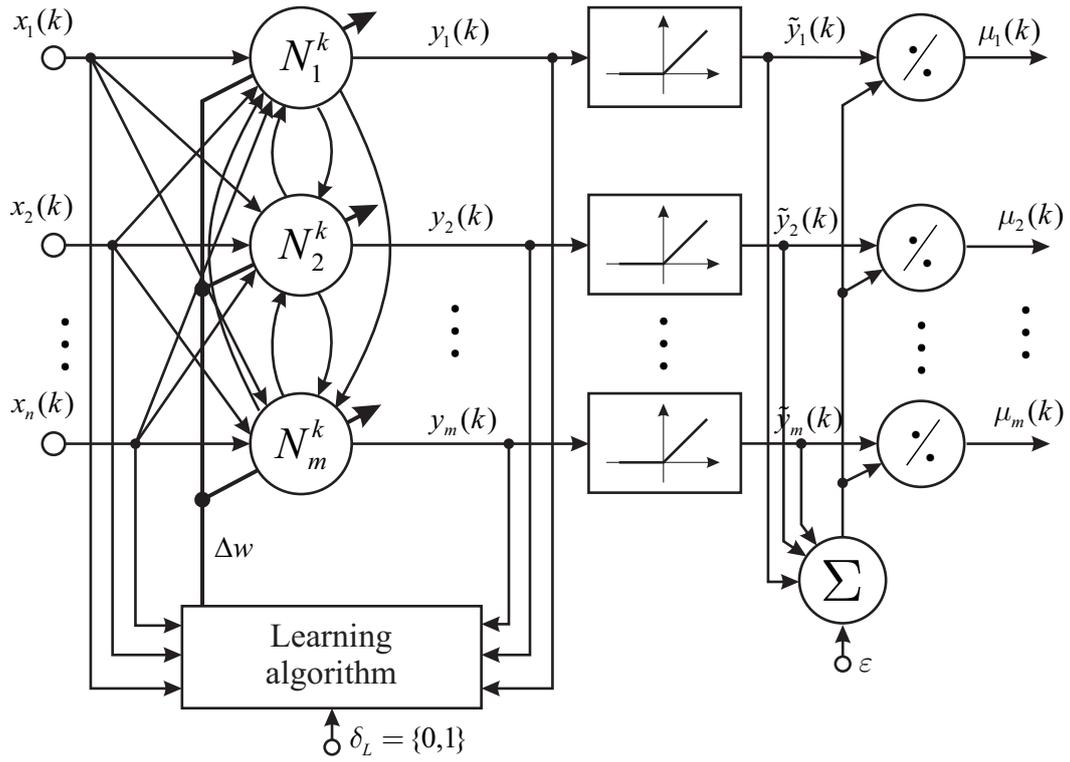

Fig. 3. Architecture of hybrid clustering-classification neural network

For many data mining problems, especially in healthcare domain, substantial shares of input samples lack clear class association. Indeed, a diagnosis might be either missing, be ambiguous or yet unknown.

In this case it is possible to tune the synaptic weights with unified architecture with lateral connections using different learning methods. Each learning method is initialized according to the level of apriori information about $x(k)$ available.

Fig. 3 shows the architecture of hybrid clustering-classification neural network.

A resulting learning algorithm for combined (SOM+LVQ) neural network can be presented in the following form [18]:

$$w_j(k+1) = \begin{cases} \dfrac{w_j^*(k) + \eta(k)(x(k) - w_j^*(k))}{\|w_j^*(k) + \eta(k)(x(k) - w_j^*(k))\|}, \\[2mm] \delta_L(k)\dfrac{w_j^*(k) - \eta(k)(x(k) - w_j^*(k))}{\|w_j^*(k) - \eta(k)(x(k) - w_j^*(k))\|}, \\[2mm] \delta_L = \begin{cases} 1, & \text{if the network works} \\ & \text{in supervised mode} \\ & \text{and sample } x(k) \text{ does} \\ & \text{not belong to class } j, \\ 0, & \text{otherwise,} \end{cases} \\[2mm] w_j(k) \text{ for neurons, which did not} \\ \text{get activated by sample } x(k). \end{cases} \quad (34)$$

where $w_j^*(k)$ is winning neuron.

It is clear, that for $\delta_L = 0$, i.e. in self-learning mode, the first expression of formula (34) can be replaced by expression (24).

VI. RESULTS OF EXPERIMET

Reactive arthritis (ReA) is the important medical-social problem of today's world because of high prevalence of ReA among people of different age groups. ReA occupies one of the top position among inflammatory disease of joints [19-21]. In different countries the frequency of ReA equal to from 8% to 41%.

According to some studies, ReA is the result of overproduction of proinflammatory cytokines or under ReA Th1-immune response is reduced in favor of Th2-immune response [22, 23]. There are scientific studies that prove the participation of cytokines in the pathogenesis of rheumatic diseases, the most common confirming the hypothesis of the pathogenesis of ReA, which is based on an imbalance of cytokines [24].

However, by now many researcher can't establish reasons, which the infection process by the causative agent one in some individuals does not lead to medical problem, while others lead to the progress of infectious and inflammatory diseases in the acute or chronic form.

The study involved 150 children, including immunology research in 40 pediatric patients with ReA in acute form and after 9-12 months since the dawn of the disease.

Immunological research includes the study of measure of cellular, humoral, monocyte-phagocytic components of immune system, the content of cytokines (IL-1β, IL-6). In cellular component of immune system the lymphocyte subpopulations was determined (CD3+, CD4+, CD8+, CD25, CD21). Determination of serum immunoglobulin of classes A, M, G is performed by spectrophotometry.

Monocyte-phagocytic component of immune system was evaluated based on the phagocytic and metabolic activity of leukocytes by determining the phagocytic number, spontaneous and stimulated NBT-test. For estimation of cytokine status in the examined patients the levels of IL-1β and IL-6 in serum is determined based on "sandwich"-method ELISA (enzyme linked immunosorbent assay) using monoclonal antibodies.

Therefore the data set is presented by the table "object-properties" consists of 40 objects and 12 properties. The hybrid clustering-classification neural network with learning algorithm (34) was used for solving of diagnosis task of ReA based on immunological indicators of pediatric patients. The initial values of synaptic weights were taken in a random way.

As the criterion of clustering-classification we used the percent of incorrect classified patterns in test sample.

The fig. 4 shows the results of clustering-classification of data which describes immunological indicators of pediatric patients with ReA.

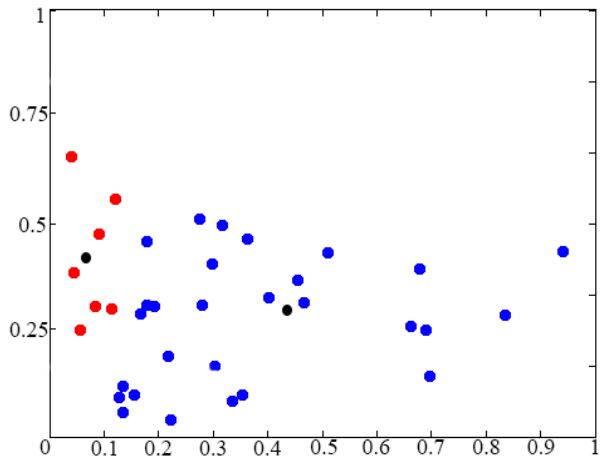

Fig. 4. Results of clustering-classification data based on hybrid clustering-classification network (● – the cluster with prolonged, recrudescent and chronic form of ReA disease, ● – the cluster with recovering patients after treatment, ● – the prototypes of clusters)

Table 1 shows the comparative analysis of clustering-classification results based on four approaches: proposed network with its learning algorithms, Kohonen self-organized map and LVQ-neural network with conventional learning algorithm and fuzzy C-means clustering algorithm.

Analysis of clustering-classification results shows the forming of two cluster, which are characterized two group of pediatric patients: group with prolonged, recrudescent and chronic form of ReA disease and group with recovering patients after proposed treatment.

The acute period of ReA in pediatric patients is characterized by the changes of immunological homeostasis in form reducing of the relative content of CD8, CD25 in serum and increasing levels of CD21, IL-6, the phagocytic number and spontaneous NBT test.

Table 1. The comparative analysis of clustering-classification results

| Neural network / Learning methods | Error of clustering-classification, train set | Error of clustering-classification, check set |
|---|---|---|
| Hybrid clustering-classification neural network / Proposed learning algorithm | 3.1 % | 4.5 % |
| Kohonen self-organized map / Conventional learning algorithm | 6.9 % | 8.0 % |
| LVQ neural network / Conventional learning algorithm | 7.9 % | 8.5 % |
| Fuzzy C-means clustering algorithm | 6,8% | 7,4% |

Therefore, it is apparent that proposed approach provides best results of clustering-classification among considered approaches due to operation in learning/self-learning mode under missing of diagnosis data labeling or in the case when the diagnosis was not made because of assident diseases of patients.

Medical data mining result was diagnosis task solving for defining of hidden factors of reactive arthritis progression in the children, which allowed for appropriate treatment.

VII. CONCLUSIONS

The combined self-learning procedure for Kohonen neural network is proposed. Such method allows data processing under the overlapping classes condition, when memberships of training data to specific classes can be unknown at all, and have both crisp and fuzzy nature. This method is based on using similarity measure, recurrent optimization and fuzzy inference and differs with high speed, possibility of operating in on-line mode and simplicity of computational realization. Medical data mining result was diagnosis task solving for defining of hidden factors of reactive arthritis progression in the children, which allowed for appropriate treatment.